\documentclass[submission,copyright,creativecommons]{eptcs}
\providecommand{\event}{ASQAP 2025} 

\usepackage{iftex}
\usepackage{graphicx}%
\usepackage{multirow}%
\usepackage{amsmath,amssymb,amsfonts}%
\usepackage{amsthm}%
\usepackage{mathrsfs}%
\usepackage[title]{appendix}%
\usepackage{xcolor}%
\usepackage{textcomp}%
\usepackage{manyfoot}%
\usepackage{booktabs}%
\usepackage{algorithm}%
\usepackage{algorithmicx}%
\usepackage{algpseudocode}%
\usepackage{listings}%

\usepackage{subcaption}
\usepackage{tikz}
\usetikzlibrary{arrows}
\usetikzlibrary{shapes.misc}
\usetikzlibrary{shapes.geometric}
\usetikzlibrary{fit}
\usetikzlibrary{backgrounds}

\usepackage{orcidlink}
\usepackage{pgfplots,pgfplotstable}
\usetikzlibrary{pgfplots.statistics}
\usepgfplotslibrary{external} 
\tikzsetexternalprefix{tikzfig/}
\usepackage[utf8]{inputenc}
\usepgfplotslibrary{groupplots}

\ifpdf
  \usepackage{underscore}         
  \usepackage[T1]{fontenc}        
\else
  \usepackage{breakurl}           
\fi

\title{A Case Study on the Application of Digital Twins for Enhancing CPS Operations}

\author{Irina Muntean \orcidlink{0009-0004-2407-9304}
\institute{TYTAN Technologies\footnote{This author was affiliated with fortiss GmbH, Munich, Germany at the time of conducting this research.}\\ Munich, Germany}
\email{irina@tytan-technologies.com}
\and
Mirgita Frasheri \orcidlink{0000-0001-7852-4582}
\institute{DIGIT, Dept. of Elec. and Comp. Engineering\\
Aarhus University, Denmark}
\email{\quad mirgita.frasheri@ece.au.dk}
\and
Tiziano Munaro \orcidlink{0009-0007-1959-7803}
\institute{fortiss GmbH\\ Munich, Germany}
\email{\quad munaro@fortiss.org}
}

\def\titlerunning{Digital Twins for Enhancing CPS Operations}
\def\authorrunning{I. Muntean, M. Frasheri, T. Munaro}
\begin{document}
\maketitle

\begin{abstract}
To ensure the availability and reduce the downtime of complex cyber-physical systems across different domains, e.g., agriculture and manufacturing, fault tolerance mechanisms are implemented which are complex in both their development and operation. 
In addition, cyber-physical systems are often confronted with limited hardware resources or are legacy systems, both often hindering the addition of new functionalities directly on the onboard hardware.
Digital Twins can be adopted to offload expensive computations, as well as providing support through fault tolerance mechanisms, thus decreasing costs and operational downtime of cyber-physical systems.
In this paper, we show the feasibility of a Digital Twin used for enhancing cyber-physical system operations, specifically through functional augmentation and increased fault tolerance, in an industry-oriented use case.
\end{abstract}

\section{Introduction}\label{sec:introduction}

Cyber-physical systems (CPSs) pose ongoing challenges for engineers and testers due to their increased complexity, scale, and autonomy~\cite{Leite2019,Ebert2016}, necessitating an adaption of development and deployment processes.
On the other hand, Digital Twins (DTs), as virtual replicas of CPSs~\cite{kritzinger2018digital}, also referred to  as Physical Twins (PTs), emerge as a powerful tool to monitor and provide a wide variety of services to their PTs, such as fault diagnosis, reconfiguration etc.~\cite{Douthwaite2021}, across diverse domains such as agriculture, healthcare, industrial automation, and logistics.
In line with agile development processes, DTs have the potential to be instrumental in developing CPS stepwise, from fully digital models to symbiotic Hardware-in-the-Loop simulations to the fully implemented robotic system consisting of both PT and DT.
This would allow for iterative validation and testing, as well as run-time augmentatation of the CPSs functionalities through DT services.
Due to limited hardware resources or legacy systems, it is not always possible to implenent enhanced functionalities directly on the board hardware~\cite{Ebert2016}.
%
%
By offloading new functionalities to a DT—whether on-premise or cloud-based—organizations can reduce acquisition and operational costs without compromising system capabilities, crucial for avoiding financial costs, especially in industrial settings.
As CPSs require fault tolerance mechanisms which are complex in both their development and operation, providing non-safety-critical redundancy as a DT service could be beneficial for ensuring durability and availability.

The \textbf{contribution} of this work is an empirical evaluation of how a DT can enhance system capabilities and fault tolerance in an industry-oriented robotic platform.
Specifically, the case study shows that DTs can (1) augment physical systems by extending their functionalities, and (2) increase fault tolerance by taking over tasks from malfunctioning components, thereby reducing system downtime. 

\section{DT - PT Infrastructure}\label{sec:approach}

In this section we describe the infrastructure needed to support the enhancement of the Physical Twin (PT) operations through its Digital Twin (DT). 
In scenarios where a PT is being developed alongside its DT, the same framework could be adopted for enabling early testing by means of Hardware-in-the-Loop simulations, with the benefit of a seamless transition from development to deployment.
The infrastructure shall consist of: (i) a CPS representing the PT, (ii) a platform hosting the DT, and (iii) a communication infrastructure which allows the twins to communicate either over a network, or peer-to-peer, according to a communication protocol of choice. 
Both PT and DT platforms should support the execution of development artefacts, e.g., simulation models, software components etc.
In addition, the twins must \textbf{synchronize} to ensure they operate on correct, up-to-date data.  
This requires mechanisms for the \textbf{synchronization} of the components executed on both twins in parallel, ensuring that the all the involved executable subsystems progress or "step" forward in time in a synchronized manner.
This can be achieved in multiple ways: 
(i) the DT is responsible for the synchronization, sending stepping commands to the PT, 
(ii) the PT is responsible for the synchronization, sending stepping commands to the DT,
and (iii) both DT and PT step independently and sync with each other at given intervals.
The choice between these alternatives is use-case dependent.

\section{Case Study}\label{sec:experiment}
In order to assess the applicability of a Digital Twin(DT)-based approach for Physical Twin (PT) enhancement, we instantiate a generic infrastructure on the \textit{fortissimo} rover platform. 
This robotic platform serves as the PT, and is an industry-oriented use case involving Advanced Driver-Assistance Systems (ADASs) and Automated Driving (AD) functions.
The case study will capture two different test scenarios.
The first experiment demonstrates the ability of a DT to augment the capabilities of the PT, whereas in the second experiment, the DT increases the fault tolerance of the system and reduces the down-time by taking over the functionality of the PT's faulty components.

    \subsection{The fortissimo rover}
The \emph{fortissimo} rovers 
are customized scale 1:10 model cars equipped with redundant sensors, namely two ultrasonic sensors, one laser sensor and one camera, and a distributed E/E platform architecture. They serve as a teaching and evaluation platform for model-based systems engineering (e.g., \cite{Dantas2022}) at the fortiss Mobility Lab\footnotemark \footnotetext{https://www.fortiss.org/en/research/fortiss-labs/detail/mobility-lab}. The rovers feature a highway pilot: an autonomous driving function capable of taking over both longitudinal and lateral control within a specific operational domain. Its functionality is modeled in AutoFOCUS3\footnotemark \footnotetext{https://www.fortiss.org/en/results/software/autofocus-3}, a model-based tool supporting the design, development and validation of safety-critical embedded systems. 
The rover's DT was co-developed as a co-simulation of its behavior within a customized CARLA simulation~\cite{Dosovitskiy17} of its environment.


\subsection{Infrastructure}\label{subsec:sysinfra}
For demonstrating the broad applicability of the presented concept across different configurations and simulation environments in the realm of cyber-physical systems, we propose an experimental testbed~\footnote{A replication package can be made available upon request.} encompassing two distinct co-simulation frameworks building upon the Functional Mock-up Interface (FMI) standard: ROSCo\footnotemark \footnotetext{https://git.fortiss.org/ff1/rosco} and Maestro~\cite{thulebuilding}. 
The Maestro co-orchestration engine is used on the DT side, whereas the ROSCo framework is applied on the PT side.
According to the FMI standard, Functional Mock-up Units (FMUs) are used to specify the behavior of the software units involved and can be exported from various tools, including among others, MATLAB Simulink, OpenModelica, IBM Rational Rhapsody, CATIA and AutoFOCUS3. 

For our experiments, we instantiate the generic infrastructure with the given components. 
The DT is hosted on enterprise hardware, an Intel\textsuperscript{®} Core™ i7 Notebook with 8 cores, 16 GB RAM and Ubuntu 20.04., and can run any FMI-compliant co-simulation framework, in this case the Maestro based co-simulation.
The corresponding PT is the \textit{fortissimo} rover equipped with a Raspberry Pi 3 Model B, redundant, multi-modal front-facing sensors: Two ultrasonic sensors, a LIDAR, and an RGB camera. 
The PT will execute the ROSCo co-simulation, as ROSCo is part of the \textit{fortissimo} platform and can access and control the hardware of the vehicle.
The communication infrastructure between the twins builds upon the Advanced Message Queuing Protocol (AMQP) protocol, discussed below.
Moreover, both tools support fault injection, enabling us to conduct the aforementioned experiments by assessing the behavior before and after fallback mechanisms intervene.

\textbf{ROSCo and Maestro Co-simulations:}
The two co-orchestration frameworks ROSCo and Maestro handle the joint execution of FMUs in a co-simulation context. 
The ROSCo co-simulation framework introduced in \cite{Munaro2022} builds upon ROS~\cite{Thomas2014} and the FMI standard to construct a lightweight co-simulation architecture.
ROSCo takes the given FMUs as input, wraps them into ROS nodes, co-simulates these by stepping them synchronously using ROS services, and manages the data flow among them, enabling the communication of FMUs during the co-simulation.
The framework also enables the injection of various types of hardware and software faults which have been introduced in ~\cite{Munaro2024} and accurately simulates their propagation throughout the system. 
The Maestro co-orchestration engine was introduced in~\cite{thulebuilding} and provides flexible, configurable co-simulation time progression, as well as fault-injection support~\cite{frasheri2021fault,Frasheri&23a}.  
Here, FMUs are stepped in parallel, with input/output exchanges taking place after all FMUs have stepped.

\textbf{Communication:}\label{subsec:communication}
The chosen communication infrastructure uses the AMQP protocol - an open-standard protocol designed for messaging between applications in a reliable and efficient manner. RabbitMQ\footnotemark \footnotetext{https://www.rabbitmq.com/} is an open-source message broker software that implements the AMQP protocol and serves as an intermediary between producers (applications that send messages) and consumers (applications that receive messages).
The PT and DT communicate through a RabbitMQ server by sending and receiving messages in JSON format.
Therefore, both ROSCo and Maestro have interfaces for communicating with AMQP and dealing with the incoming data based on a user-defined configuration encoding the input-output connections between the PT and DT.

\textbf{Maestro-AMQP Interface:}
On the DT side, a RabbitMQ data broker is encapsulated within an FMU and thereafter used as a bridge to external components.
This component is named RMQFMU~\cite{Thule2020a}, and enables a user to configure its inputs and outputs in accordance with the data to be sent and received. 
Inputs define data sent out externally via the RabbitMQ server, while outputs determine data received and forwarded to the other FMUs.
The supported data types for inputs and outputs are Real, Integer, Boolean, and String. 

\textbf{ROSCo-AMQP Interface:}
The ROSCo framework was extended to incorporate a RabbitMQ interface for the PT. 
Specifically, a ROS node encapsulating a RabbitMQ publisher and subscriber is used for orchestrating the data transmission and reception via the RabbitMQ server and therefore, enabling the communication between the twins. 

\textbf{ROSCo based Fault injection:}
For the following experiments, we will use the fault injection framework ﬁrst presented in \cite{Munaro2022} built in ROSCo, enabling the simulation of \textit{Loss of Software Unit} on both PT and DT, for validating the fallback mechanisms before deploying the proposed PT and DT artefacts in a real-world setting.
The loss of such a software unit is replicated by terminating the respective ROS node. 
Any processing and communication is hereby instantly halted.


\subsection{Experiments}
Two experiments were selected to illustrate the capability of our DT-enabled infrastructure to facilitate (1) {functional augmentation} and (2) {increased fault tolerance} on the DT. 

\textbf{Setup:}
Both PT and DT execute the whole \textit{fortissimo} rover model illustrated by the light blue components in Fig.~\ref{fig:taskarchitecture}. 
Besides the mutual components, the PT and DT execute additional functionality and corresponding fallback mechanisms either for functional augmentation or for increased fault tolerance. 
Fig.~\ref{fig:taskarchitecture} depicts the software architecture present in Experiment 1, where the dark blue components represent functional augmentation and corresponding fallback mechanism.
The twins initiate the FMI-based co-simulation of the rover model in a synchronized manner.
The PT sends all the environment inputs at every step, including sensor values and manual user inputs to the DT, so they can simulate in parallel, provided the same inputs.
The execution of the whole system on both platforms given the same environment inputs is advantageous during early validation activities, such as conformance testing, where the co-developed DT can provide support in fault localization acting as an oracle.
This, however, is beyond the scope of this work.
For the purpose of this paper, the synchronization is such that, when available, the DT controls the stepping of the whole system.
Otherwise, should the DT be unreachable, e.g. due to a communication breakdown, the PT either continues its intended operation if a fallback is available on the PT, or switches to a defined safe mode until the communication with the DT is re-instated.
Moreover, in order to verify the availability of the twins, a mutual heartbeat mechanism is integrated into both twins --- as the twins exchange heartbeats at each step, these signals are also used for synchronization, triggering the stepping of the FMUs on each side.


\textbf{Experiment 1: Augmenting PT capabilities}
The first experiment demonstrates the ability of a DT to reduce the costs of acquisition and operation while augmenting CPS capabilities by proposing an advanced Adaptive Cruise Control (ACC) feature. 
Moreover, it provides an example of the DT's capabilities at run-time.
For the purpose of this experiment, we employ two versions for the ACC feature and make the following distinction between a basic and an advanced ACC: 
The basic ACC, represented by \texttt{ACC Fallback}, uses solely laser sensor values to regulate the rover speed by assessing the distance to the front obstacles, 
whereas the advanced ACC, corresponding to DT's \texttt{ACC Augmented}, fuses the laser and ultrasonic sensor data for considering obstacles also on its left and right sides.
To assess the feasibility of the augmentation, we apply fault injection on the DT-side simulating loss of communication due to, e.g. network failures, and analyze the corresponding PT response, specifically we inject a fault into DT's \texttt{Heartbeat} component.
We place obstacles in the front, left, and right sides of the rover before and after the DT becomes unavailable to observe the behavior of the advanced and basic ACC features.

\textbf{Experiment 2: Increasing fault tolerance}
The second experiment illustrates the DT's ability to enhance fault tolerance and reduce the downtime of the CPS by taking over functionality that the PT can no longer offer reliably. 
In contrast to the previous experiment, the ACC provided by the DT is a replica of the basic ACC executed primarily on the PT. 
The goal is to ensure that in the event of ACC instability or failure on the PT, the DT's ACC seamlessly takes charge, guaranteeing continuous system operation despite faults, 
thereby increasing the fault tolerance through redundancy.

To demonstrate the increased fault tolerance, we inject a fault into the PT's ACC unit.
We place obstacles in front of the rover, both before and after the ACC on the PT-side stops operating to observe the fail-operational behavior provided by the DT.  
The experiment also shows the utility of the approach for conducting tests as close to the actual operation as possible, given that the DT-PT setup is the same.

\subsection{Test Results}
The experimental results obtained using the \textit{fortissimo} rover case-study are presented in the following paragraphs, demonstrating the feasibility of the proposed approach.
\begin{figure}[!htpb]
\begin{subfigure}[b]{0.5\textwidth}
  \centering
  \includegraphics[width=1.0\textwidth]{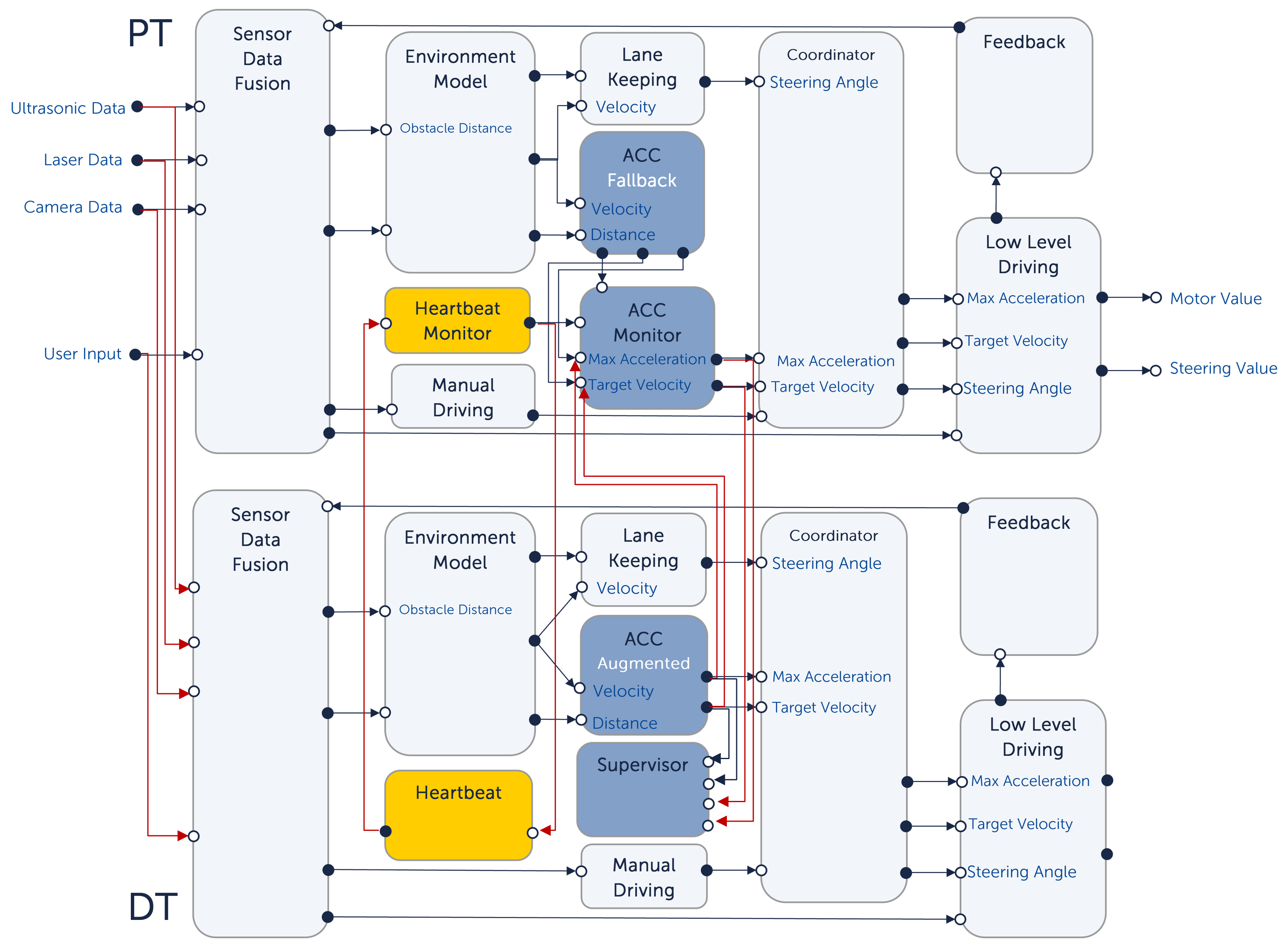}
  
  \caption{}\label{fig:taskarchitecture}
  \end{subfigure}
  \begin{subfigure}[b]{0.5\textwidth}
        \centering
      \resizebox{1.0\textwidth}{!}
      {
%
%
\definecolor{mycolor1}{rgb}{0.92941,0.69412,0.12549}%
\definecolor{mycolor2}{rgb}{0.63529,0.07843,0.18431}%
\definecolor{mycolor3}{rgb}{0.00000,0.44706,0.74118}%

\def\transformtime#1 #2:#3:#4!{
    \pgfkeys{/pgf/fpu=true,/pgf/fpu/output format=fixed}%
    \count2=#2 \edef\hour{\the\count2}%
    \count2=#3 \edef\min{\the\count2}%
    \count2=#4 \edef\sec{\the\count2}%
    \pgfmathparse{\hour*3600-\pgfkeysvalueof{/pgfplots/timeplot zero}*3600+\min*60+\sec + 0}%
    \pgfkeys{/pgf/fpu=false}%
}

\pgfplotsset{
    timeplot zero/.initial=0,
    timeplot/.style={
        x coord trafo/.code={\expandafter\transformtime##1!},
        x coord inv trafo/.code={%
            \pgfkeys{/pgf/fpu=true,/pgf/fpu/output format=fixed}
            \pgfmathsetmacro\hours{floor(##1/3600)+\pgfkeysvalueof{/pgfplots/timeplot zero}}
            \pgfmathsetmacro\minutes{floor((##1-(\hours-\pgfkeysvalueof{/pgfplots/timeplot zero})*3600)/60)}
            \pgfmathsetmacro\seconds{##1-floor(##1/60)*60}
             \def\pgfmathresult{\pgfmathparse{mod(\hours,60)<10?"0":{},int(mod(\hours,60))}\pgfmathresult:\pgfmathparse{mod(\minutes,60)<10?"0":{},int(mod(\minutes,60))}\pgfmathresult:\pgfmathparse{mod(\seconds,60)<10?"0":{},int(mod(\seconds,60))}\pgfmathresult}
            \pgfkeys{/pgf/fpu=false}
        },
    scaled x ticks=false,
    xticklabel=\tick,
    }
}

\pgfplotstableread[col sep=comma,header=true]{experiment-augmenting-capabilities.csv}\data

\pgfplotstableread[col sep=comma]{experiment-increased-fault-tolerance.csv}\dataa%

\begin{tikzpicture}

\begin{groupplot}[
    group style={
        group size = 1 by 3,
        group name=myplotsa,
        vertical sep=80pt,
        horizontal sep=0pt,
        ylabels at=edge left,
        yticklabels at=edge left,
    },
    height=1.5cm,
    width=8cm,
    xlabel near ticks,
    ylabel near ticks,
    grid,
    grid style={dashed,gray!40},
    axis line style={gray!40},  
    axis on top=false,
scale only axis,
ylabel style={rotate=0},
legend style={at={(0.0,1.8)}, draw=none, fill=none, anchor=north west, legend cell align=left, align=left,},
]
\nextgroupplot[%
ylabel={Heartbeat \#},
timeplot,
x tick label style={align=center, rotate=30}, 
]
\addplot[color=mycolor1,smooth] table[col sep=comma,x index=0,y index=1] {\data};
\addlegendentry{PT Heartbeat}
\nextgroupplot[%
ylabel={Target Acceleration},
        timeplot,
x tick label style={align=center, rotate=30}, 
]
\addplot[color=mycolor2,smooth] table[col sep=comma,x index=0,y index=3] {\data};
\addlegendentry{PT Target Acceleration}
\addplot[color=mycolor3,dashed] table[col sep=comma,x index=0,y index=5] {\data};
\addlegendentry{DT Target Acceleration}
\nextgroupplot[%
ylabel={Target Velocity},
xlabel={Time},
        timeplot,
x tick label style={align=center, rotate=30}, 
]
\addplot[color=mycolor2,smooth] table[col sep=comma,x index=0,y index=2] {\data};
\addlegendentry{PT Target Velocity}
\addplot[color=mycolor3,dashed] table[col sep=comma,x index=0,y index=4] {\data};
\addlegendentry{DT Target Velocity}

\end{groupplot}

 \node[text width=6cm,align=center,anchor=south] at ($(myplotsa c1r1.south)+(0,-2)$) {\label{}};
 \node[text width=6cm,align=center,anchor=south] at ($(myplotsa c1r2.south)+(0,-2)$) {\label{}};
 \node[text width=6cm,align=center,anchor=south] at ($(myplotsa c1r3.south)+(0,-2.5)$) {\label{}};
 
 \end{tikzpicture}
 
 \begin{tikzpicture}%
 \begin{groupplot}[
    group style={
        group size = 1 by 3,
        group name=myplots,
        vertical sep=80pt,
        horizontal sep=0pt,
        ylabels at=edge left,
        yticklabels at=edge left,
    },
    height=1.5cm,
    width=8cm,
    xlabel near ticks,
    ylabel near ticks,
    grid,
    grid style={dashed,gray!40},
    axis line style={gray!40},  
    axis on top=false,
scale only axis,
ylabel style={rotate=0},
legend style={at={(0.0,1.8)}, draw=none, fill=none, anchor=north west, legend cell align=left, align=left,},
]
\nextgroupplot[%
ylabel={Heartbeat \#},
    timeplot, 
x tick label style={align=center, rotate=30}, 
]
\addplot[color=mycolor1,smooth] table[col sep=comma,x index=0, y index=1] {\dataa};
\addlegendentry{PT Heartbeat}
\nextgroupplot[%
ylabel={Target Acceleration},
    timeplot,
x tick label style={align=center, rotate=30}, 
]
\addplot[color=mycolor2,smooth] table[col sep=comma,x index=0, y index=3] {\dataa};
\addlegendentry{PT Target Acceleration}
\addplot[color=mycolor3,dashed] table[col sep=comma,x index=0, y index=5] {\dataa};
\addlegendentry{DT Target Acceleration}
\nextgroupplot[%
ylabel={Target Velocity},
xlabel={Time},
    timeplot, 
x tick label style={align=center, rotate=30}, 
]
\addplot[color=mycolor2,smooth] table[col sep=comma,x index=0, y index=2] {\dataa};
\addlegendentry{PT Target Velocity}
\addplot[color=mycolor3,dashed] table[col sep=comma,x index=0, y index=4] {\dataa};
\addlegendentry{DT Target Velocity}

\end{groupplot}

 \node[text width=6cm,align=center,anchor=south] at ($(myplots c1r1.south)+(0,-2)$) {\label{}};
 \node[text width=6cm,align=center,anchor=south] at ($(myplots c1r2.south)+(0,-2)$) {\label{}};
 \node[text width=6cm,align=center,anchor=south] at ($(myplots c1r3.south)+(0,-2.5)$) {\label{}};
 
\end{tikzpicture}%
}
    \caption{}\label{fig:exp1}
      \end{subfigure}
      \caption{(a) Example PT and DT models generated from the \textit{fortissimo} rover development model for {functional augmentation}, i.e. for Experiment 1. The red arrows represent the the input-output connections between the PT and DT.
  The \texttt{Supervisor} component provides visualisation of the PT's and DT's ACC behavior to operators.
  (b) \textit{Left side, Exp. 1:} Rover drives, ACC is activated at 08:38:46. DT heartbeat received on the PT (top). Target Acceleration values from the ACC on the DT and PT respectively (middle). Target Velocity values from the ACC on the DT and PT respectively (bottom). Until 08:38:46 the PT is controlled by the DT (dashed blue line). After, the PT takes over (red line). \textit{Right side, Exp. 2:} Rover drives, ACC is activated at 09:03:04. At 09:05:04 the PT ACC is killed and DT ACC takes over. DT heartbeat received on the PT (top). Target Acceleration values from the ACC on the DT and PT respectively (middle). Target Velocity values from the ACC on the DT and PT respectively (bottom). Before 09:05:50, the PT uses its own ACC (red line). After, it is controlled by the DT (blue line).}
\end{figure}
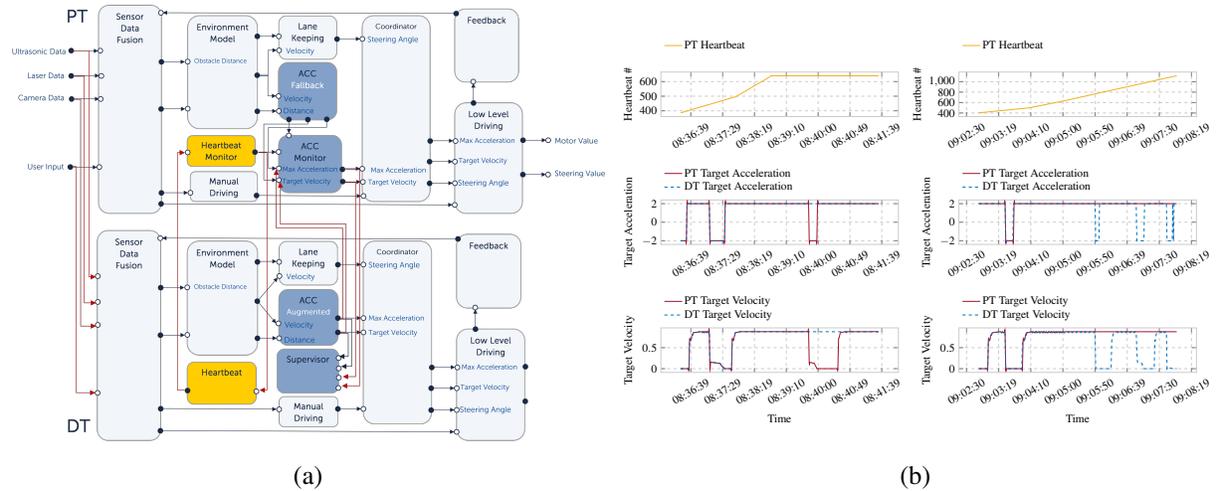

\textbf{Experiment 1: Augmenting PT Capabilities}
The DT and the rover are initiated, and soon after, the \texttt{Advanced ACC} feature is activated manually (at 08:36:38).
As there are no obstacles in front of the vehicle, the ACC increases the Target Acceleration and Velocity (as illustrated in Fig. \ref{fig:exp1}, left column), causing the rover to start driving.
The functionality of the advanced ACC provided by the DT is tested first (at 08:37:10), by placing obstacles in the front, right, and left of the rover. 
Irrespective of where the obstacles are put, it is observed that the rover stops due to the ACC, which uses both laser and ultrasonic sensor values.
This is shown by the decreasing Target Acceleration and Velocity values, which are set to their minimum values -2 and 0, respectively.
The obstacles are removed at 08:37:45, and the rover starts driving again. 
In order to make the DT unavailable for the purposes of this test, fault injection is applied into 
DT's \texttt{Heartbeat} unit at 08:38:46, resulting in communication breakdown between the twins.
It can be observed in Fig.~\ref{fig:exp1} (left column) that the heartbeat is no longer incremented afterwards. 
After the fault injection, PT's basic ACC, \texttt{ACC Fallback}, takes over, as \texttt{ACC Monitor} detects DT's unavailability. 
It is observed that the rover stops in case of front obstacles, however not for obstacles placed to the left or right, as the basic ACC solely uses the laser sensor values.
After the front obstacles are removed, the rover starts driving again (at 08:40:32).
Therefore, we can conclude that the DT-approach eliminates the need for additional onboard hardware and therefore reduces acquisition and operational costs.

\textbf{Experiment 2: Increased fault tolerance}
The DT and rover are initiated. 
PT's ACC is activated manually at 09:03:04, resulting in increased Target Acceleration and Velocity determining the vehicle to start driving, as there are no obstacles detected.
Thereafter, an obstacle is placed in front of the rover (at 09:03:31).
It is observed that the car stops.
With the removal of the obstacle at 09:03:57, the rover starts driving again.
At 09:05:04, fault injection is applied into PT's ACC leading to the termination of the respective ROS node.
This occurrence is detected by the monitor placed on the PT, which starts considering the ACC values sent by the ACC replica deployed on the DT.
In Fig.~\ref{fig:exp1} (right column) it can be observed, that after 09:05:50, the values generated by the DT (dashed blue line) are what control the PT.
After an obstacle is placed in front of the rover at 09:05:51, it stops, due to the DT.
The rover starts driving at 09:06:16 after the obstacle has been removed.

\section{Discussion and Concluding Remarks}\label{sec:conclusion}

DT applications span across different sectors, with a heavy focus on industrial manufacturing, related to pump, engine and battery production~\cite{Somers2023}.
Architectural patterns for PT-DT integration have been explored in various contexts, including~\cite{Gil2024} and ~\cite{Tekinerdogan20}, providing a foundation for these applications.
DTs can offer a wide range of services, such as detecting failures through real-time monitoring, or degradation monitoring, thus learning about the life-cycle of different components~\cite{Somers2023}, e.g., in battery life-cycle management~\cite{Eaty2023}.
More advanced functionalities such as failure anticipation or more generally prediction~\cite{Piardi2020} are also of interest, as well as predictive maintenance which aims to estimate when certain parts of the PT should be changed/upgraded.
Other possible services include product customisation and optimisation, and sub-domains, namely separation processes, advanced robotics, and additive manufacturing~\cite{Bottjer2023}.
In regards to collaborative robotics, Douthwaite \textit{et. al} have proposed a modular DT framework targeting the safety assurance of industrial collaborative robotics processes~\cite{Douthwaite2021}.

The contribution in this paper lies in a case study showcasing the applicability of a DT for enhancing cyber-physical systems operations. 
Specifically, we showed (i) how a DT can augment the capabilities of its PT running remotely on-premise, not bound by resource and computation constraints, while also allowing for fallback mechanisms should the DT become unavailable, and (ii) how the DT can provide increased fault tolerance by taking over certain functionalities in case of failures in the PT.

The main limitation in this paper concerns the realism of the adopted case study of an autonomous rover, especially the functionality under study, i.e. the Adaptive Cruise Control (ACC).
However, in domains such as agriculture, where the speeds of the robots would be limited, as well as traffic and the interaction with humans, the ACC example is representative.
Another limitation relates to the assumption of best-effort real-time communication, without provision of guarantees. 

Future work will investigate the adequacy of the proposed approach for safety-critical research, as well as its adaption for real-time communication, where formal methods could be adopted to provide guarantees. 
In addition, we are interested in investigating in-depth a methodology for DT-PT co-development, that fits well the DevOps gap in CPS development, a promising venue of research~\cite{Semeraro2021}. 
%



\bibliographystyle{eptcs} 
\bibliography{generic}
\end{document}